\documentclass[runningheads]{llncs}

\usepackage{eccv}

\usepackage{eccvabbrv}

\usepackage{graphicx}
\usepackage{booktabs}
\usepackage{multirow}
\usepackage{amsmath}
\usepackage{upgreek}
\usepackage[normalem]{ulem}
\usepackage[accsupp]{axessibility}  % Improves PDF readability for those with disabilities.

\usepackage[pagebackref,breaklinks,colorlinks,citecolor=eccvblue]{hyperref}
\usepackage{orcidlink}

\begin{document}

% ---------------------------------------------------------------
% TODO REVIEW: Replace with your title
\title{Leveraging Thermal Modality to Enhance Reconstruction in Low-Light Conditions} 

% TODO REVIEW: If the paper title is too long for the running head, you can set
% an abbreviated paper title here. If not, comment out.
\titlerunning{Thermal-NeRF}

% TODO FINAL: Replace with your author list. 
% Include the authors' OCRID for the camera-ready version, if at all possible.
\author{Jiacong Xu\inst{1} \and
Mingqian Liao\inst{2} \and
K Ram Prabhakar\inst{1} \and
Vishal M. Patel\inst{1}}

% TODO FINAL: Replace with an abbreviated list of authors.
\authorrunning{Xu et al.}
% First names are abbreviated in the running head.
% If there are more than two authors, 'et al.' is used.

% TODO FINAL: Replace with your institution list.
\institute{Johns Hopkins University, Baltimore MD 21218, USA \\
\email{\{jxu155,rprabha3,vpatel36\}@jhu.edu} \and
Southern University of Science and Technology, Shenzhen, P.R.China \\
\email{12012919@mail.sustech.edu.cn}}

\maketitle

\begin{abstract}
  Neural Radiance Fields (NeRF) accomplishes photo-realistic novel view synthesis by learning the implicit volumetric representation of a scene from multi-view images, which faithfully convey the colorimetric information. However, sensor noises will contaminate low-value pixel signals, and the lossy camera image signal processor will further remove near-zero intensities in extremely dark situations, deteriorating the synthesis performance. Existing approaches reconstruct low-light scenes from raw images but struggle to recover texture and boundary details in dark regions. Additionally, they are unsuitable for high-speed models relying on explicit representations. To address these issues, we present Thermal-NeRF, which takes thermal and visible raw images as inputs, considering the thermal camera is robust to the illumination variation and raw images preserve any possible clues in the dark, to accomplish visible and thermal view synthesis simultaneously. Also, the first multi-view thermal and visible dataset (MVTV) is established to support the research on multimodal NeRF. Thermal-NeRF achieves the best trade-off between detail preservation and noise smoothing and provides better synthesis performance than previous work. Finally, we demonstrate that both modalities are beneficial to each other in 3D reconstruction. Code and dataset can be accessed via \url{https://github.com/XuJiacong/Thermal-NeRF}
  \keywords{Neural Radiance Field \and Low-light Enhancement \and Thermal Imaging \and Novel View Synthesis \and Multimodality}
\end{abstract}

\section{Introduction}
\label{sec:intro}
Neural Radiance Field (NeRF) \cite{nerf} is a technique that renders high-quality images of 3D scenes for novel viewpoints by learning the neural volumetric representation of the geometry and lighting. The compelling novel view synthesis performance of NeRF has inspired many research efforts \cite{mipnerf360, zipnerf, 3dgaussiansplatting, instantngp, blocknerf, neus}.

\begin{figure}[t!]
\centering
    \includegraphics[width=\textwidth]{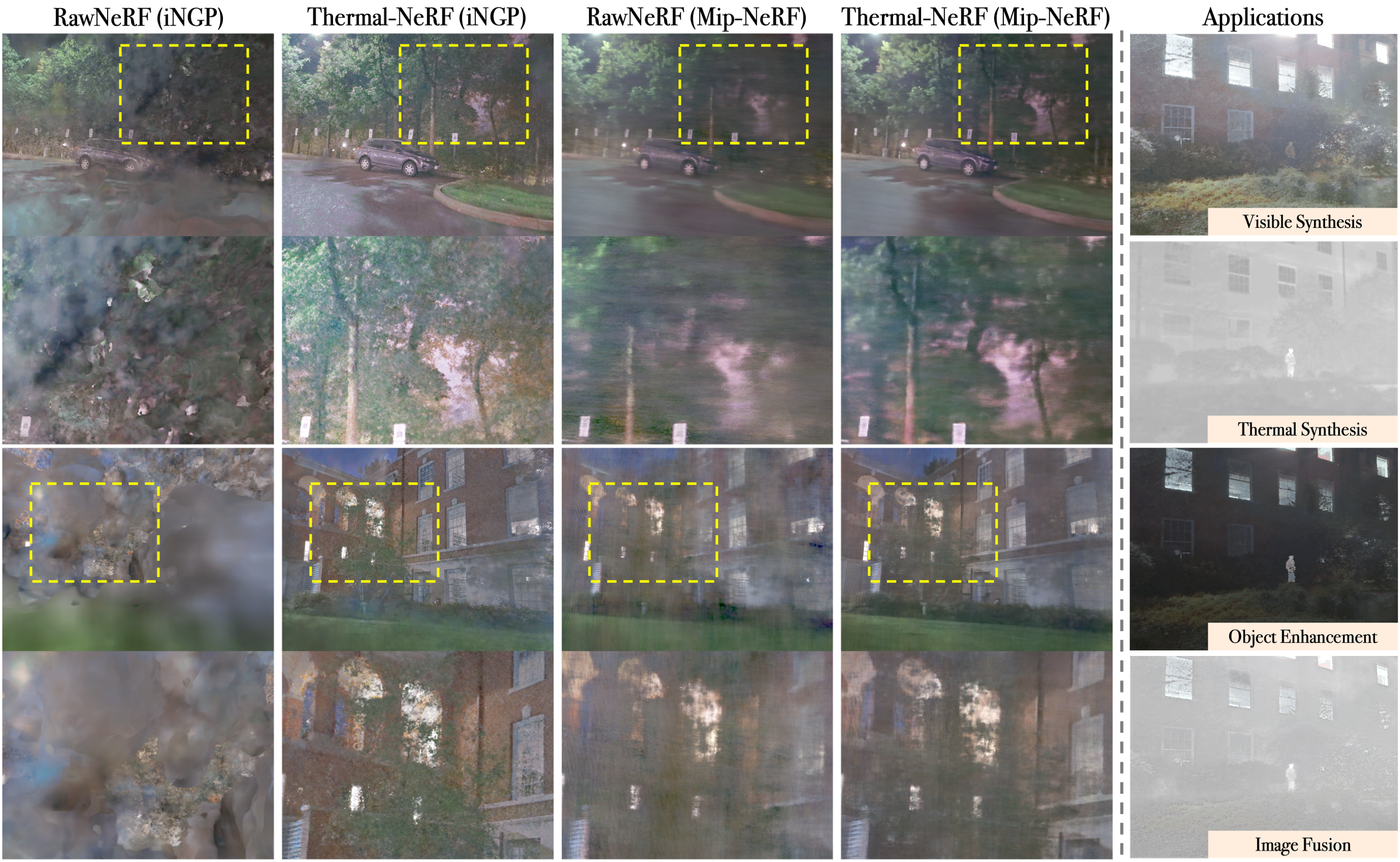}
\caption{Illustration of the synthesis performances of RawNeRF \cite{rawnerf} and our proposed Thermal-NeRF and its multimodal applications. We implement RawNeRF and Thermal-NeRF on Mip-NeRF \cite{mipnerf} and iNGP \cite{instantngp}. The first four columns present the qualitative comparison of raw image rendering (postprocessed) of different methods.  With the raw and thermal images generated by Thermal-NeRF, we can adjust the visible environment and highlight the thermal objects or fuse the two modalities by image fusion \cite{swinfusion} for better visual effects in some applications.}
\vspace{-7
 mm}
\label{fig:intro}
\end{figure}

NeRF requires faithful colorimetric radiance to parse the 3D information from high-quality camera images. However, in low-light conditions where most of the pixel intensities are noisy and close to zero, insufficient illumination tends to blur or distort dark pixels, thus reducing synthesis performance \cite{llnerf, alethnerf}. Previous works combine the 2D low-light image enhancement methods with NeRF models to simultaneously optimize scene representation and enhancement parameters, presenting vivid color and details for dark scenes using postprocesssed 8-bit sRGB images. Notably, the lossy camera ISP and compression algorithms limit the dynamic range of the images and remove valuable details in the dark regions, resulting in blocks of zero values. This limits geometry and shading details recoverable by the current synthesis-and/then-enhancement methods.

Instead of sRGB images, RawNeRF \cite{rawnerf} modifies the NeRF model to learn the 3D scene representations from linear raw images that preserve the original camera sensor responses in 16-bit pixel intensities, and shows the flexibility of the rendered raw images for different visual effects, such as exposure adjustment, tonemapping, and defocus. Even though the MLP-based NeRF model \cite{mipnerf} used by RawNeRF can smooth the noises in the raw images, the detailed boundary and texture information are also blurred in the extremely dark regions (Figure \ref{fig:intro}). Besides, when transferred to fast models relying on or partially relying on explicit representations \cite{instantngp, 3dgaussiansplatting, plenoxels}, RawNeRF fails to reconstruct the scene and sometimes cannot converge to the geometry. To leverage any possible information in dark regions, we follow RawNeRF and focus on the view synthesis of raw images for visible scene reconstruction in low-light situations.
%in extremely dark situations.

Thermal camera, which is insensitive and robust to illumination variations, weather conditions, and other extreme conditions \cite{vtuav}, can highlight targets with temperatures different from environment \cite{llvip}, such as human, fire, and mobile vehicles. It has been widely equipped in modern drone, surveillance, autonomous vehicle, and robot systems, serving as a complementary sensor for object detection \& tracking \cite{detection, tracking}, fire monitoring \cite{fire}, building inspection \cite{building}, and etc. Given the strength of thermal imaging in low-light scenarios, we incorporate thermal modality into 3D scene reconstruction and establish the first multi-view thermal \& visible dataset (MVTV) to study the effect of multimodal images on novel view synthesis and support the research on multimodal NeRF.

In most real-world scenarios, capturing long-exposure or multi-exposure images or videos is impractical, especially for hand-held and mobile devices, such as drones, or in emergent situations. Therefore, our objective is to reconstruct the 3D scene with short-exposure visible and thermal images and accomplish visible and thermal view synthesis simultaneously in low-light conditions. With the rendered raw visible and thermal images, we can adjust the visible environment freely and highlight the thermal objects for better visual effect (Figure \ref{fig:intro}), which can be applied in many scenarios, such as 3D localization of rescue or surveillance \cite{rescue}. We provide implementations of our method on Mip-NeRF \cite{mipnerf} and iNGP \cite{instantngp} for a fair comparison with previous works. Also, our method can be seamlessly transferred to 3DGS \cite{3dgaussiansplatting} to obtain a better and faster solution, which is discussed in the supplementary materials.

The main contributions of this paper are three-fold:
\begin{itemize}
  \item We introduce Thermal-NeRF to enhance the scene reconstruction in low-light conditions using thermal and visible images, and it achieves better rendering performance compared with previous methods.
  \item We demonstrate that our method resolves the stability issue of previous works and thereby can be seamlessly transferred to explicit or hybrid models for faster rendering speed and comparable rendition quality.
  \item We establish the first aligned multi-view thermal \& visible dataset (MVTV), which consists of 20 night scenes (10 of them contain long-exposure ground truth), to facilitate the research on multimodal NeRF.
\end{itemize}
\section{Related Work}
\subsection{Neural Radiance Field}
NeRF \cite{nerf} parses the 3D scene via implicit representation and accomplishes photo-realistic novel view synthesis, which significantly outperforms previous techniques, such as SRNs \cite{srns}, Local Field Light Fusion \cite{locallightfieldfusion}, Neural Volumes \cite{neuralvolumes}, and earlier classic methods \cite{earlier1, earlier2}. Recent advanced works based on NeRF architecture further improved the synthesis performance on dealing with aliasing \cite{mipnerf}, unbounded scene \cite{mipnerf360},  unstructured images \cite{nerfw}, sparse view \cite{pixelnerf}, glossy surfaces \cite{refnerf}, HDR radiance field \cite{hdrnerf}, and etc. In addition to RGB color radiance, depth signal \cite{dsnerf} and point cloud \cite{pointnerf}, which can be automatically obtained from SFM, are leveraged for fewer training views and faster training speed. 

\begin{figure}
\centering
    \includegraphics[width=0.8\textwidth]{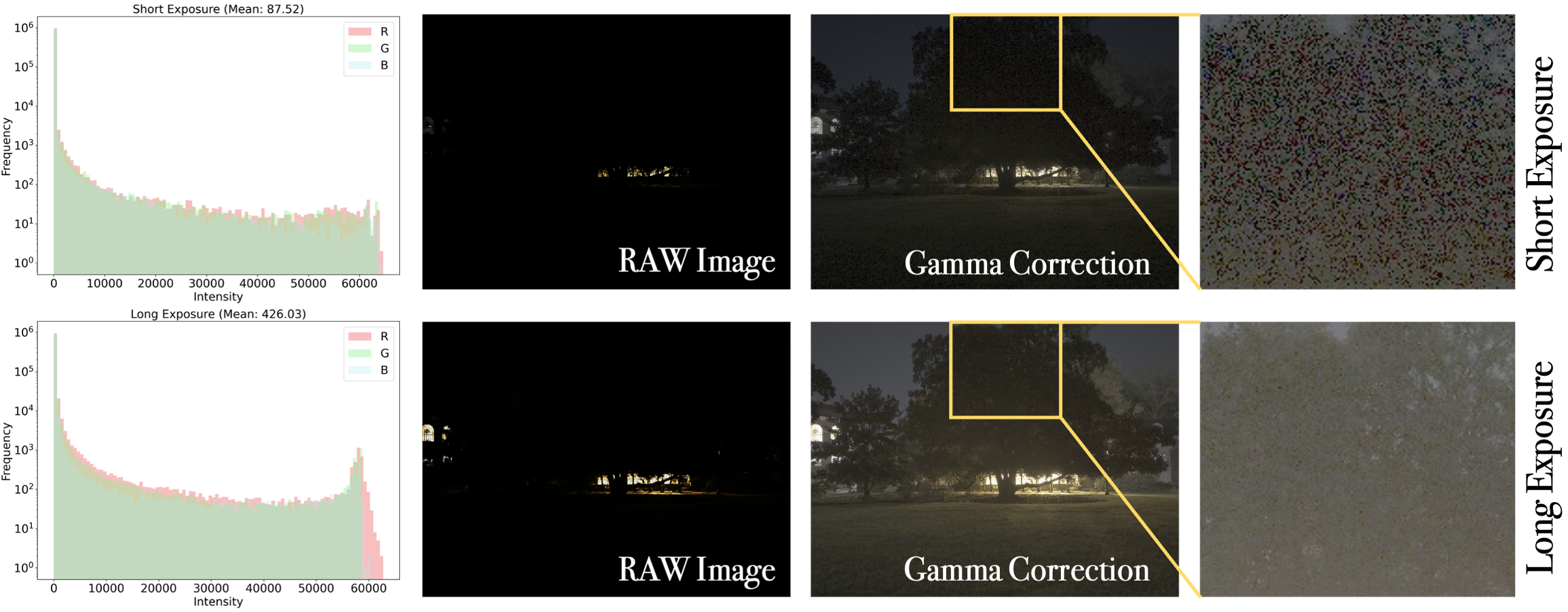}
\caption{Pixel intensity distributions of raw images taken by short and long exposures. The noise level in dark regions is very high, and long exposure will alleviate this issue.}
\label{fig:hist}
\end{figure}

For robust view synthesis, researchers are seeking solutions to apply NeRF on images captured in various adversarial conditions, such as deblurring \cite{deblurnerf}, denoising \cite{nan}, and camera pose refining \cite{sparf}. Similar to this work, the research of reflectance \& lighting \cite{nerv, nerd, nerfactor, neilf} decouples the density and appearance of the original NeRF into geometry (surface normal), material (BRDF), and lighting for better flexibility on editing and relighting. These complex graphic modeling methods are not designed for low-light scenes, and relighting is not required in our scenarios, so inverse rendering based on NeRF is not involved.

The training and inference speeds of vanilla NeRF are the major obstacles that prevent it from being used in real-world applications, so researchers introduced many strategies to speed up NeRF. Instead of representing the scene by implicit volume, explicit voxel grid, tree, or 3D Gaussian \cite{nsvf, plenoctrees, plenoxels,baking, dvgo, 3dgaussiansplatting} are utilized to constrain sampling or interpolate learnable parameters for computational efficiency at the cost of memory usage. The tensor radiance field \cite{tensorf}, which can be factorized into compact low-rank components, is leveraged to achieve a memory-efficient representation of voxel grid features. iNGP \cite{instantngp} presents fast training speed and high-quality reconstruction by concatenating the interpolations of multi-resolution hash grid values for positional encoding. Recently, 3D Gaussian Splatting \cite{3dgaussiansplatting} utilizes controllable 3D Gaussians to represent the scene and directly splats Gaussians to the screen for real-time rendering.

\subsection{NeRF in Low-light Condition}
The colorimetric radiances of a scene convey the geometry and shading information, which are extracted and preserved by the NeRF learning models. However, when the illumination of the scene is not enough, such as in night scenarios, the radiance reaching the camera is low and will be contaminated by the built-in sensor noises. Besides, the lossy camera ISP will further reduce the contained information by a series of non-linear operations, which results in zero values in the dark regions of camera-finished sRGB images (8-bit). LLNeRF \cite{llnerf} follows the Retinex strategy, which has been widely applied in the enhancement of 2D low-illumination images \cite{deepretinex, uretinex, retinex3, retinex4,lime}, decomposes the radiance of the point into reflectance and illumination and manipulates the illumination using a learned gamma correction process, which outperforms the combination of existing enhancement methods and NeRF. Aleth-NeRF \cite{alethnerf} attaches a concealing field to suppress radiance when training the model on low-light images and removes the concealing field when synthesizing the novel views.

However, when the scene is extremely dark, the low-amplitude signals will be eliminated by the camera ISP process or image compression algorithms (JPEG), which leads to the situation that there is no information to recover and enhance. Then, existing sRGB-based methods become inapplicable (detailed analysis in supplementary materials). Raw image preserves the radiance signals received by the camera sensor to the most significant extent, and the 16-bit color intensity can capture an extensive range of sensor responses. Thus, RawNeRF \cite{rawnerf} directly reconstructs near-dark scenes from raw images and demonstrates that the MLP in NeRF will automatically smooth the zero-mean noises contained in the raw images. In our experiments, we observe that RawNeRF fails to reconstruct the detailed geometries and textures in highly dark situations and is not suitable for fast models relying on explicit representations \cite{instantngp, 3dgaussiansplatting}. Explicit or hybrid models are sensitive to the geometry constraints, while the RawNeRF strategy cannot faithfully extract the geometry information from dark raw images with highly imbalanced intensity distributions (Figure \ref{fig:hist}).

\subsection{Multimodal NeRF}
Due to the lack of available datasets specifically established for novel view synthesis with more modalities other than RGB images, the research on multimodal NeRF is relatively inadequate. 
Zhu \textit{et al.} \cite{multimodalnerf} present the first work using multiple modalities (RGB, Infrared, Point cloud) as input for the same NeRF model and attempt to address the alignment issue for different modalities since the multi-sensor images in the employed navigation dataset, M2DGR \cite{m2dgr}, are not strictly paired and aligned. Chang \textit{et al.} \cite{lidar} select several sequences from the self-driving dataset Argoverse 2 \cite{argoverse} and put strong 3D geometry priors for ray sampling of NeRF using LiDAR maps. Poggi \textit{et al.} \cite{xnerf} introduce learnable poses and normalized cross-devices coordinate to tackle the spatial alignment of multi-spectral images (visible, NIR). To expedite the research on multimodal NeRF and enable researchers to focus on method development instead of image alignment, we establish the first aligned multi-view thermal and visible dataset and demonstrate that both modalities benefit each other in 3D low-light scene reconstruction.
\begin{figure}[t]
\centering
    \includegraphics[width=\textwidth]{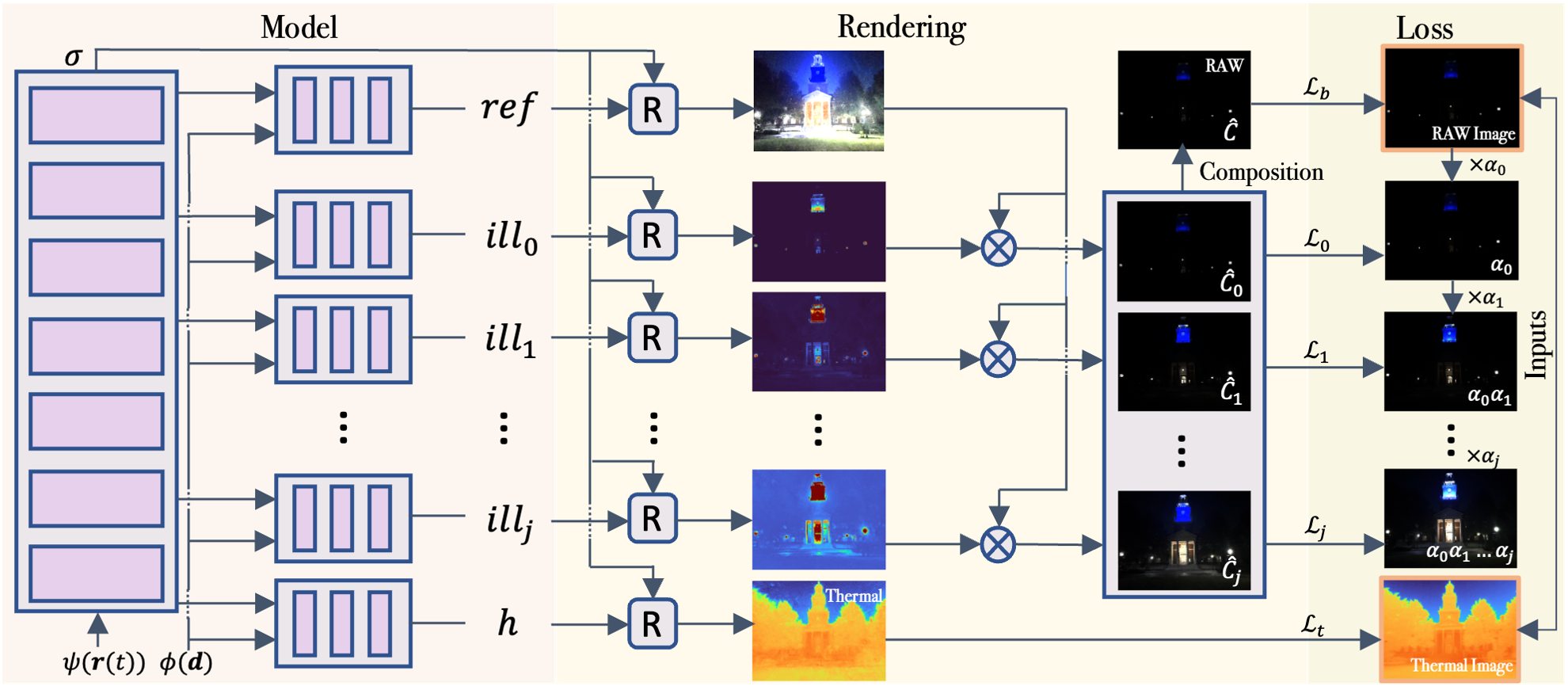}
\caption{Overview of the architecture of Thermal-NeRF. $ref$, $ill_j$, and $h$ represent the reflectance, $j$-th illumination, and thermal radiance, respectively. R refers to volume rendering. The model part can be adjusted based on different scene representations.}
\label{fig:model}
\end{figure}
\section{Method}
\label{sec:method}
In this section, we introduce Thermal-NeRF, which consists of 3 major components: \textit{i) Thermal Enhancement} leverages thermal supervision to constrain the optimization of the geometry; \textit{ii) Retinex3D} updates the Retinex \cite{retinex3} decomposition and transfers it from 2D images to 3D scenes; \textit{iii) Multi-Exposure} manipulates the illuminations to render images targeting different ranges of intensity and composites the rendered images recursively to generate the final raw images. The entire pipeline is illustrated in Figure \ref{fig:model}.
\subsection{NeRF Preliminary}
NeRF \cite{nerf} represents the volume density for every point in the scene and describes the radiance field by predicting the three-dimensional color given viewing directions for these points, which is written as:
\begin{equation}
    F_{\Theta}(\textbf{x}, \theta, \phi)=(\textbf{c}, \sigma)
\end{equation}
where $F_{\Theta}$ is a trainable Multi-Layer Perceptron (MLP), $\textbf{x}$ refers to the 3D coordinates of the point in the space, $\theta$ and $\phi$ define the azimuthal and polar angles for the viewing direction, $\textbf{c}$ and $\sigma$ represent the predicted radiance and volume density. Each pixel in the image will define a camera ray, $\textbf{r}(t)=\textbf{o}+t\textbf{d}$, using camera position $\textbf{o}$ and viewing direction $\textbf{d}$. Then the color $C(\textbf{r})$ for that pixel can be rendered using volume rendering \cite{volume_rendering} by the predicted $\textbf{c}$ and $\sigma$:
\begin{equation}
\label{eq:render_rgb}
    C(\textbf{r})=\int_{t_n}^{t_f}T(t)\sigma(\textbf{r}(t))\textbf{c}(\textbf{r}(t), \textbf{d})dt,
\end{equation}
where $T(t)$ represents the accumulated transmittance from camera $\textbf{r}(t_n)$ to the point $\textbf{r}(t)$ and can be calculated by:
\begin{equation}
    T(t)=exp\Bigl(-\int_{t_n}^{t}\sigma(\textbf{r}(s))ds\Bigr).
\end{equation}
By uniformly sampling the points along the ray, the integration can be computed by numerical approximation:
\begin{equation}
\label{eq:rgb_numerical}
    \hat{C}(\textbf{r})=\sum^{N}_{i=1}\alpha_{i}T_{i}c_{i}\text{, where } T_{i}=exp(-\sum^{i-1}_{j=1}\sigma_{j}\delta_{j}),
\end{equation}
where
$\alpha_{i}$ is the opacity of point $i$ and $\alpha_{i}=1-exp(\sigma_{i}\delta_{i})$. $\delta_{i}$ refers to the distance between adjacent sample points.

\subsection{Thermal Enhancement}
Since dark regions of an image count less compared to bright regions with large intensity in normal MSE loss, simply optimizing the network with imbalanced RGB information will lead the model to focus on brightness reconstruction instead of geometry. NeRF++ \cite{nerf++} suggests that a good geometry will remedy the shape-radiance ambiguity and enhance the synthesis performance, especially for the MLP-less and hybrid models \cite{plenoxels, dvgo, nsvf, instantngp}, which utilize explicit scene representations to store the 3D geometric information. Even though the balanced loss proposed by RawNeRF \cite{rawnerf} can slightly alleviate this issue by intensity normalization, when transferring it to iNGP \cite{instantngp} or 3DGS \cite{3dgaussiansplatting}, we observe that the model attempts to fit the color of training images by modifying the shape instead of the radiance, which will deteriorate generalization ability of the model and cause the final synthesis results to be unrealistic.

We argue that the failure of RawNeRF originates from the fact that the models with tiny MLP or without MLP cannot regress the complicated intensity distribution (Figure \ref{fig:hist}) of dark raw images, so they will force their explicit elements to spread into the empty areas to fit the low intensities, which occupy more pixels compared to large intensities. Thermal images can recover the geometric information of a scene in extremely dark situations. Thus, we provide an extra prediction head for thermal view synthesis to constrain the geometry of the scene using the same rendering equation \ref{eq:rgb_numerical}, which is represented as:
\begin{equation}
    \hat{H}(\textbf{r})=\sum^{N}_{i=1}\alpha_{i}T_{i}h_{i},
\end{equation}
%\begin{equation}
%    H(\textbf{r})=\int_{t_n}^{t_f}T(t)\sigma(\textbf{r}(t))h(\textbf{r}(t), \textbf{d})dt
%\end{equation}
where $h_{i}$ is the thermal radiance for sample point $\textbf{r}(t_{i})$, with the corresponding thermal scene reconstruction loss:
\begin{equation}
    \mathcal{L}_{t}(\hat{H}(\textbf{r}), H(\textbf{r}))=\Vert \hat{H}(\textbf{r}), H(\textbf{r}) \Vert_2.
\end{equation}
Also, the balanced loss can be regularized by the predicted thermal intensity to emphasize the foreground objects:
\begin{equation}
    \mathcal{L}_{b}(\hat{C}(\textbf{r}), C(\textbf{r}))=\sum_{i}\Bigl(\frac{\hat{C}(\textbf{r})-C(\textbf{r})}{sg(\hat{C}(\textbf{r}))+\epsilon}\cdot sg(\hat{H}(\textbf{r}))\Bigr)^2,
\end{equation}
where $sg(*)$ is the stop-gradient operation to avoid backpropagation for thermal prediction. Besides, with the extra thermal prediction, the model can satisfy the applications requiring thermal scene reconstruction.

\subsection{Retinex3D}
\label{subsec:retinex3d}
%LLNeRF \cite{llnerf} introduces the Retinex method from the field of image enhancement to light up the NeRF trained on postprocessed images.
Retinex-based low-light image enhancement methods \cite{deepretinex, uretinex, retinex3, retinex4,lime} accomplish encouraging enhancement performance in 2D case by decomposing the pixel color of an image ($I$) into Reflectance ($R$) and Illumination $L$:
\begin{equation}
\label{eq:retinex}
    I=R\cdot L.
\end{equation}
LLNeRF \cite{llnerf} leverages this technique to light up the NeRF by manipulating the illumination $L$ and assumes that the reflectance of a point is isotropic and independent of the viewing direction. However, in the natural world, the lighting conditions and the surface materials are complex, and sometimes, the reflectance could be anisotropic when the surface material is not Lambertian.

Therefore, we feed both the directional and density embeddings to the prediction MLPs of reflectance and illumination fields, which will enable both of them to be view-dependent and lead the model to be more general. Different from previous work that applies the Retinex multiplication (\ref{eq:retinex}) in the 3D space for each point, we render the reflectance $Ref$ and illumination $Ill$ for each pixel and then multiply them directly in image space for better performance:
\begin{equation}
    Ref(\textbf{r})=\sum^{N}_{i=1}\alpha_{i}T_{i}ref_{i}, \text{ } Ill(\textbf{r})=\sum^{N}_{i=1}\alpha_{i}T_{i}ill_{i},
\end{equation}
where $ref_{i}$ and $ill_{i}$ represent the reflectance and illumination fields for the sample point $\textbf{r}(t_{i})$, respectively and
%\begin{equation}
%    Ref(\textbf{r})=\int_{t_n}^{t_f}T(t)\sigma(\textbf{r}(t))ref(\textbf{r}(t), \textbf{d})dt
%\end{equation}
%\begin{equation}
%    Ill(\textbf{r})=\int_{t_n}^{t_f}T(t)\sigma(\textbf{r}(t))ill(\textbf{r}(t), \textbf{d})dt
%\end{equation}
\begin{equation}
    \hat{C}(\textbf{r}) = Ref(\textbf{r}) \cdot Ill(\textbf{r}).
\end{equation}
Note that the accumulation will smooth the noises of the predictions \cite{rudnev2022nerf} in dark regions and thereby aid the entire optimization process. We define a general Retinex decomposition in 3D space and refer to this method as Retinex3D.

\subsection{Multi-Exposure}
\label{subsec:me}
The modified exponential activation function \cite{rawnerf} enables the model to parameterize linear radiance values in large dynamic range, but it still cannot resolve the imbalanced distribution issue of raw images in low-light conditions with only single short exposure. As Figure \ref{fig:hist} demonstrates, over $90\%$ of the pixels are with very low intensity, so how to extract the rich but noisy color and geometry information contained in these low-intensity pixels is the key question. Based on this observation, we propose a Multi-Exposure (ME) strategy to alleviate the regression overhead of a single prediction head and highlight any possible information contained in pixels with different scales of illumination.

Our solution is to split the range of intensity into several parts exponentially, then equip the model with multiple tiny MLP heads for the regression of different parts, which emphasizes the information in different scales and is beneficial to hybrid models, and finally compose the outputs recursively. Given the shared reflectance field, the illuminations can be amplified to generate images with different exposures since the space of raw signals is linear. Then, the color output for the $i$-th illumination can be calculated by:
\begin{equation}
    \hat{C}_{i}(\textbf{r}) = Ref(\textbf{r}) \cdot Ill_{i}(\textbf{r}).
\end{equation}
When the predicted value of current head falls into the range of the next head (Figure \ref{fig:model}), the output for the next level is more trustable, so we provide a smoothing linear sign function to control the possibility of choosing current output given all the output ranges that lay in $(0,1)$, which is defined as:
\begin{equation}
\label{eq:sign}
    f_{i}(\hat{C}_{i},\gamma)=
    \begin{cases}
      1.0, & \alpha_{i+1}\hat{C}_{i}\geq \gamma \\
      \alpha_{i+1}\hat{C}_{i}/\gamma,  & 0.0 \leq \alpha_{i+1}\hat{C}_{i} < \gamma  \\
      0.0, & \alpha_{i+1}\hat{C}_{i}<0.0,
    \end{cases}
\end{equation}
where $\alpha_i$ is the amplification coefficient for the $i$-th output, $\gamma$ is a hyperparameter and $\gamma \in (0,1)$. Since the predictions for all the heads are not perfect and contain errors, a simple sign function will cause color inconsistency and spot points in the final reconstruction, as shown in the supplementary file. 

Then, we can composite the final estimate by a recursive equation:
\begin{equation}
    \widetilde{C}_{i}=f_{i}\hat{C}_{i}+(1-f_{i})\widetilde{C}_{i+1}/\alpha_{i+1},
\end{equation}
with the initial condition that $\widetilde{C}_{j}=\hat{C}_{j}$, where $j$ refers to the last output. The final color estimate can be obtained by the stopping criterion that $\hat{C}(\textbf{r})=\widetilde{C}_{0}$.

To enable different illumination prediction heads to focus on their own scales of reconstruction intensity, we simulate images with different exposures by scaling the original raw image to the corresponding intensity levels and supervising each head by the simulated image. Thus, the loss function for each output is: 
\begin{equation}
    \mathcal{L}_{i}(\hat{C}_{i}(\textbf{r}), C(\textbf{r}), \mu_{i})=\Vert \mu_{i}\hat{C}_{i}(\textbf{r}) - \mu_{i}\prod \limits_{t=0}^{i}\alpha_{t}C(\textbf{r}) \Vert_{2},
\end{equation}
where $\mu_{i}$ indicates the dependency between current and former outputs and can be calculated by:
\begin{equation}
    \mu_{i}=
    \begin{cases}
      1.0, & i = 0 \text{ or } \alpha_{i}\hat{C}_{i-1} < 1.0  \\
      0.0, & \text{otherwise}.
    \end{cases}
\end{equation}
Thermal-NeRF consists of all the aforementioned strategies, so its total training loss $\mathcal{L}_{total}$ includes three parts: thermal enhancement loss $\mathcal{L}_{t}$ and $\mathcal{L}_{b}$, and multi-exposure loss $\mathcal{L}_{i}$.  The overlall loss can be represented as:
\begin{equation}
    \mathcal{L}_{total}=\lambda_{t}\mathcal{L}_{t}+\lambda_{b}\mathcal{L}_{b}+\sum_{i=0}^{j}\lambda_{i}\mathcal{L}_{i},
\end{equation}
where $\lambda_{t}$, $\lambda_{b}$, and $\lambda_{i}$, $i=0,1,...,j$ are hyperparameters and can be adjusted for better synthesis performance. The implementation of the Thermal-NeRF strategy on 3DGS \cite{3dgaussiansplatting} is slightly different (more details in the supplementary).
\section{Experimental Results}
In this section, we firstly introduce our multimodal dataset (MVTV) and present the thermal and visible view synthesis results corresponding to Thermal-NeRF and previous work. Then, we provide extended applications of our method and dataset. More detailed analysis and results are in the supplementary document. % too much line?
\begin{figure*}
\centering
    \includegraphics[width=0.95\linewidth]{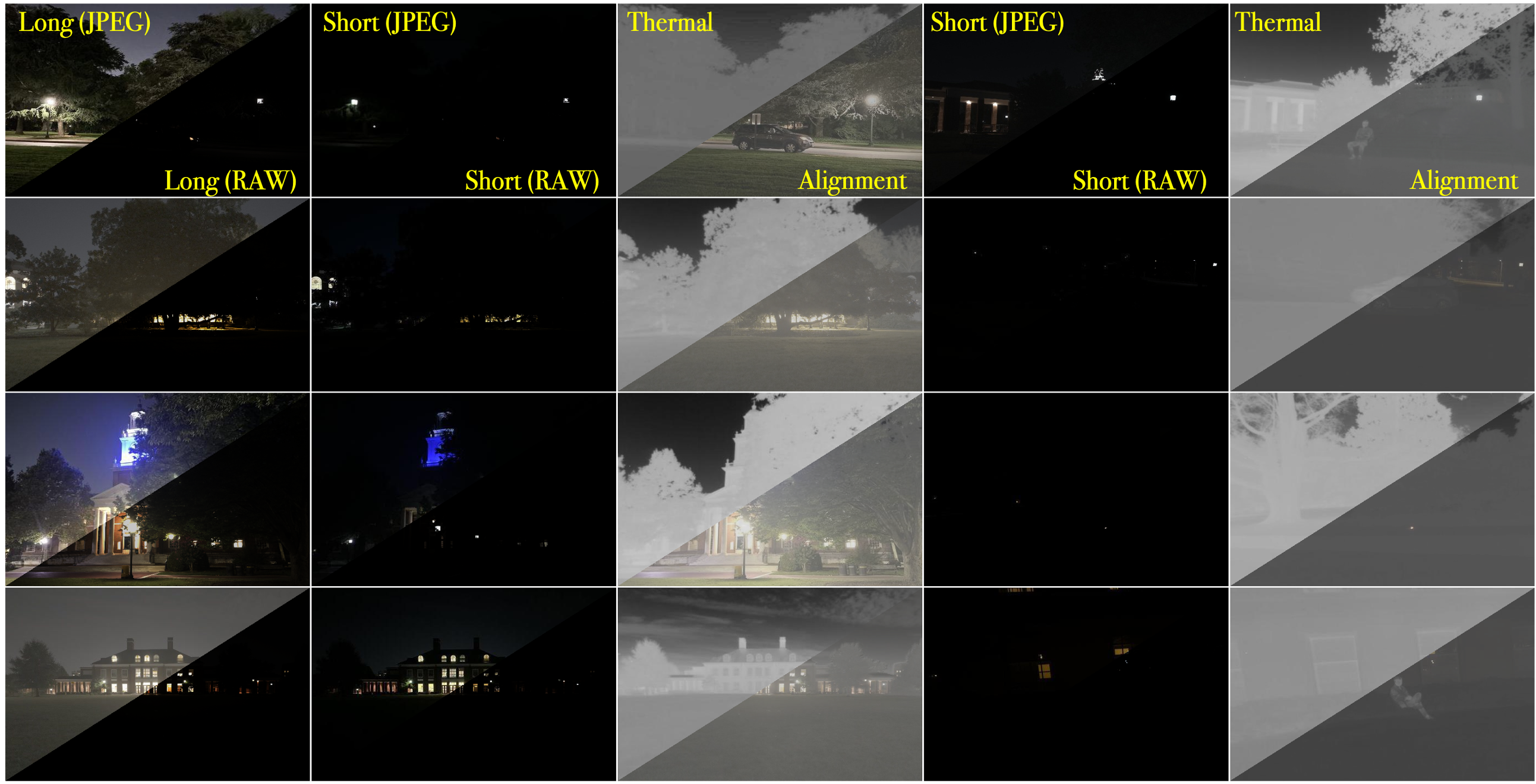}
\caption{Samples from VS (1 $\sim$ 3 columns) and TS (4 $\sim$ 5 columns). The camera poses are estimated by COLMAP \cite{colmap1, colmap2} using scaled short exposure raw images or combination with thermal images for real-world usage, where long exposure is inaccessible.}
\vspace{-7
 mm}
\label{fig:dataset}
\end{figure*}
\subsection{Dataset: MVTV}
Existing thermal \& visible datasets \cite{llvip,m3fd,vtuav,kaist} mount their cameras on fixed positions or mobile vehicles and mainly focus on the tasks of image fusion and object detection \& segmentation., so their viewing directions are limited and cannot satisfy the requirements of novel view synthesis. To study the effect of thermal imaging on night scene reconstruction, we collect the first Multi-View Thermal \& Visible (MVTV) dataset with aligned thermal and visible images, which consists of 2 parts: VS and TS. There are 10 scenes for low-light visible view synthesis with long exposure images as ground truth in VS and 10 scenes for application study and thermal reconstruction with thermal objects (human, car, fire) involved in TS. As shown in Figure \ref{fig:dataset}, each scene in VS consists of 5 files: raw and postprocessed visible images for long and short exposures and thermal image, while the TS part is collected by choosing images from videos, so it only contains the short exposure visible images and thermal images.

The thermal camera we use is FLIR VUE Pro with a fixed exposure time 12 $ms$ and working wavelength from 8 to 14 $\upmu m$. We chose the iPhone 14 Pro to be the visible camera for raw and postprocessed image collection and mounted it with the thermal camera on the same tripod to fix the rigid transformation between the two cameras. The short and long exposure options of visible images are set to be 1/64 and 1/1, and the actual exposure time for 1/1 option is over ten seconds. All the images are taken across the campus at night (8PM $\sim$ 12AM). We follow the alignment method introduced in \cite{llvip,m3fd} to align the paired thermal and visible images. Firstly, camera calibration is required to remove the distortion of thermal images. Then, we manually select 40 $\sim$ 60 corresponding points for accurate estimation of the homography matrix and apply the transformation to all the images for pixel locations of thermal images in the visible ones. Please refer to the supplementary for more statistics and establishment details.

\subsection{Implementation}
\noindent
\textbf{Instant-NGP.} For fast training and rendering speed, we build RawNeRF \cite{rawnerf} and our model on iNGP \cite{instantngp} and establish our code based on the \textit{nerfstudio} \cite{nerfstudio}. Since the tiny MLP of iNGP cannot model the highly imbalanced intensity distribution of dark raw images and large MLP will affect the backpropagation of information to grid features, we attach four parallel illumination prediction heads ($j=3$) to the model and set the amplification coefficients $\alpha_{i}=10$ for $i=1,2,3$ and $\gamma=0.9$ in the case of saturation. 

\noindent
\textbf{Mip-NeRF.} Following the default setting of RawNeRF, we also implement our model on Mip-NeRF \cite{mipnerf} to embrace the denoising power of large MLP for fair comparison. Different from iNGP, only two illumination heads ($j=1$) are required, considering that the base MLP itself is capable of parameterization of a large range of radiance values, and the extra prediction will further highlight the dark details by setting $\alpha_{1}=100$. Since only single short-exposure images are available, the variable exposure training of RawNeRF is not applicable in our scenario for HDR image rendering.

\begin{figure}
\begin{minipage}[c]{0.48\textwidth}
\centering
    \includegraphics[width=\textwidth]{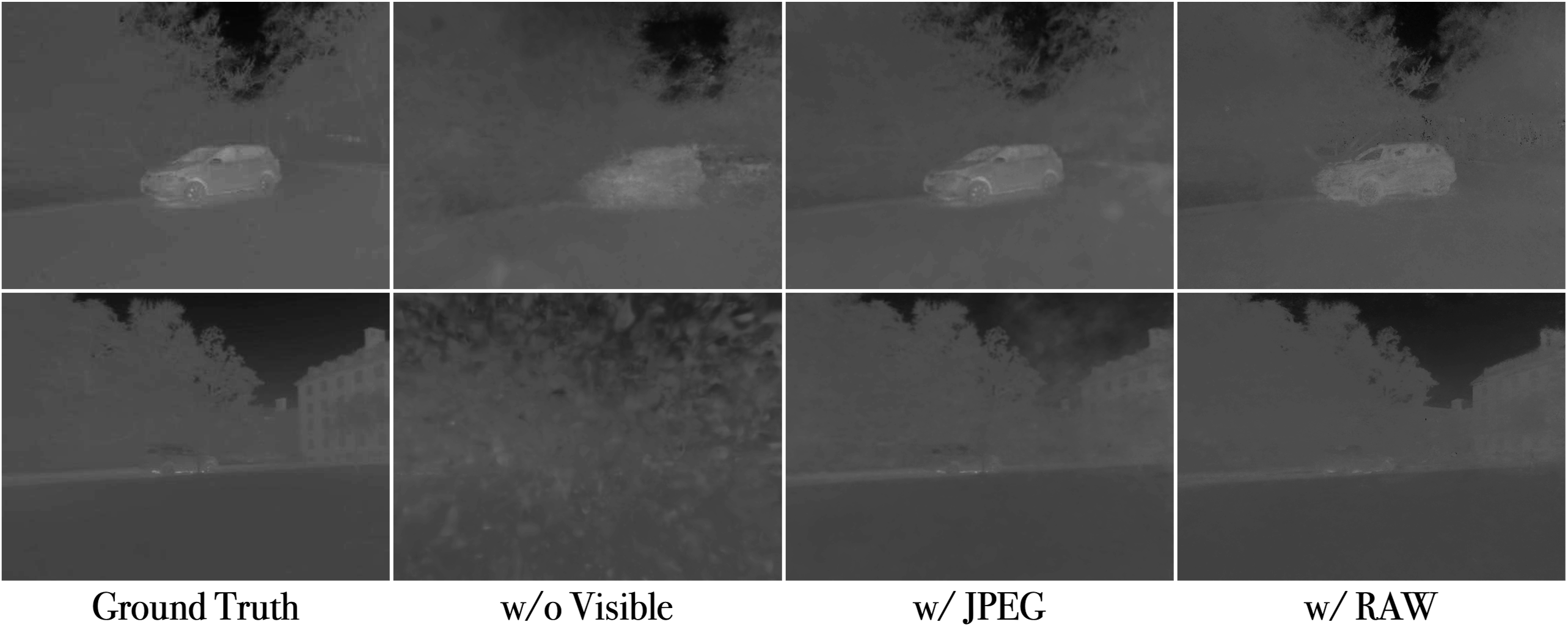}
\caption{Samples of results for thermal view synthesis. RAW and JPEG refer to using raw and post-processed images as extra supervision, respectively.
}
\label{fig:ts}
\end{minipage}
\begin{minipage}[c]{0.48\textwidth}
\centering
\begin{tabular}{c|cc|cc} 
\hline\hline
\multirow{2}{*}{Method} & \multicolumn{2}{c|}{VS-MVTV} & \multicolumn{2}{c}{TS-MVTV}  \\ 
\cline{2-5}
                        & PSNR  & SSIM            & PSNR  & SSIM            \\ 
\hline
w/o Visible             & 29.75 & 0.90           & 31.74 & 0.92          \\
w/ JPEG              & 31.75 & \textbf{0.94}           & 32.73 & \textbf{0.93}           \\
w/ RAW              & \textbf{31.79} & 0.89           & \textbf{33.06} & 0.91           \\
\hline\hline
\end{tabular}
\captionof{table}{Experimental results of Thermal-NeRF with and without visible modality for thermal view synthesis. The notations follow Figure \ref{fig:ts} for convenience. %Thermal-NeRF will degrade to iNGP after removing visible components. JPEG refers to using post-processed images as an extra supervision in iNGP.
}
\label{tab:ts}
\end{minipage}
\end{figure}

\subsection{Thermal View Synthesis}
Considering that thermal scene reconstruction is mainly required in emergent applications, we present the results of thermal view synthesis using Thermal-NeRF based on iNGP implementation for fast training and inference speed in Table \ref{tab:ts}. 
The thermal camera is sensitive to the temperature variations emitted from objects, but when the temperature difference is negligible, thermal imaging will fail to distinguish different objects and convey the depth information of the scene. Visible images, even though noisy in low-light conditions, is able to capture the geometry, texture, and lighting details in the visible spectral region, which contains rich information for 3D scene reconstruction.
We conduct experiments on both the VS and TS parts of our MVTV dataset. The results (Table \ref{tab:ts}) demonstrate that visible supervision will improve the reconstruction performance on thermal scenes. Also, the visible images can be combined with paired thermal ones by image fusion to estimate the camera poses, which are usually more accurate than thermal or visible images only. Figure \ref{fig:ts} provides the qualitative comparison of thermal scene reconstruction, which indicates that visible supervision will stabilize the training process, and Thermal-NeRF with raw visible input will further clean the floaters in the synthesis results. Empirically, it performs better than the other two options on geometry reconstruction.

\begin{table}[t]
\scriptsize
\centering
\begin{tabular}{c|ccc|ccc|ccc|ccc} 
\hline\hline
\multirow{2}{*}{Model} & \multicolumn{3}{c|}{Gilman}                    & \multicolumn{3}{c|}{Shriver}                   & \multicolumn{3}{c|}{Mason}                     & \multicolumn{3}{c}{Human}                       \\ 
\cline{2-13}
                       & PSNR           & SSIM          & LPIPS         & PSNR           & SSIM          & LPIPS         & PSNR           & SSIM          & LPIPS         & PSNR           & SSIM          & LPIPS          \\ 
\hline
RawNeRF\textsuperscript{\textdagger}                & 17.48          & 0.65          & 0.61          & 14.56          & 0.47          & 0.67          & 21.26          & 0.79          & 0.45          & 23.29          & 0.74          & 0.65           \\
RawNeRF\textsuperscript{\textdagger}+T              & 21.65          & 0.69          & 0.44          & 17.13          & 0.50          & 0.61          & 25.14          & 0.82          & 0.35          & 26.96          & 0.79          & 0.48           \\
Thermal-NeRF\textsuperscript{\textdagger}            & \uline{25.20}  & 0.73          & \uline{0.38}  & 20.16          & 0.56          & \uline{0.48}  & 25.16          & 0.82          & \uline{0.28}  & 27.29          & \uline{0.80}  & \textbf{0.39}  \\
RawNeRF                & 24.41          & \uline{0.74}  & 0.43          & \uline{21.40}  & \textbf{0.60} & 0.50          & \uline{28.72}  & \uline{0.88}  & \uline{0.28}  & \uline{27.87}  & \uline{0.80}  & 0.45           \\
Thermal-NeRF            & \textbf{25.35} & \textbf{0.77} & \textbf{0.37} & \textbf{21.60} & \textbf{0.60} & \textbf{0.46} & \textbf{30.46} & \textbf{0.89} & \textbf{0.27} & \textbf{28.24} & \textbf{0.81} & \uline{0.41}   \\ 
\hline
\multirow{2}{*}{Model} & \multicolumn{3}{c|}{Corner}                    & \multicolumn{3}{c|}{Tree}                      & \multicolumn{3}{c|}{Road}                      & \multicolumn{3}{c}{Parking}                       \\ 
\cline{2-13}
                       & PSNR           & SSIM          & LPIPS         & PSNR           & SSIM          & LPIPS         & PSNR           & SSIM          & LPIPS         & PSNR           & SSIM          & LPIPS          \\ 
\hline
RawNeRF\textsuperscript{\textdagger}                & 21.94          & 0.72          & 0.68          & 20.45          & 0.64          & 0.74          & 22.07          & 0.78          & 0.59          & 25.35          & 0.66          & 0.53           \\
RawNeRF\textsuperscript{\textdagger}+T              & 22.27          & 0.74          & 0.70          & 22.13          & 0.65          & 0.60          & 28.18          & 0.83          & 0.43          & 26.07          & 0.72          & 0.56           \\
Thermal-NeRF\textsuperscript{\textdagger}           & \uline{26.98}  & \uline{0.78}  & \textbf{0.34} & 23.76          & \uline{0.68}  & \textbf{0.49} & 29.45          & 0.83          & \textbf{0.23} & \textbf{29.78}          & 0.74          & \textbf{0.26}  \\
RawNeRF                & 24.66          & 0.75          & 0.50          & \uline{23.78}  & 0.64          & 0.53          & \uline{30.63}  & \uline{0.85}  & 0.39          & 28.91  & \textbf{0.77} & 0.43           \\
Thermal-NeRF            & \textbf{28.03} & \textbf{0.81} & \uline{0.37}  & \textbf{25.25} & \textbf{0.72} & \uline{0.50}  & \textbf{31.68} & \textbf{0.87} & \uline{0.29}  & \uline{29.42} & \textbf{0.77} & \uline{0.41}   \\
\hline\hline
\end{tabular}
\caption{Experimental results of RawNeRF \cite{rawnerf} and our model: Thermal-NeRF for visible view synthesis on VS part of MVTV. We choose PSNR, SSIM, and LPIPS$\downarrow$ as the evaluation metrics. The models marked with \textdagger are implemented on iNGP \cite{instantngp}, otherwise Mip-NeRF \cite{mipnerf}. T refers to our thermal enhancement strategy. The best and second best results are in bold and underlined, respectively.}
\vspace{-7
 mm}
\label{tab:vs}
\end{table}

\subsection{Visible View Synthesis}
Our objective for visible view synthesis is to generate clean raw visible images from noisy short exposure inputs without losing detailed information in dark regions. With the rendered raw images, we can freely adjust the postprocessing pipeline to achieve different visual effects and even accomplish HDR imaging \cite{rawhdr}. In the MVTV dataset, we take the long-exposure images as the ground truth only for evaluation, considering that long exposure will smooth the noise and capture more trustful information in dark situations. Note that current evaluation metrics, such as PSNR, SSIM, and LPIPS, are not suitable for quantification of the reconstruction performance of dark raw images, where most of the pixels are with near-zero intensity. Thus, we postprocess the rendered images and adjust the exposure to the same level of long-exposure ground truth for evaluation.

\begin{figure}[t]
\centering
\begin{minipage}[t]{0.48\textwidth}
    \includegraphics[width=\textwidth]{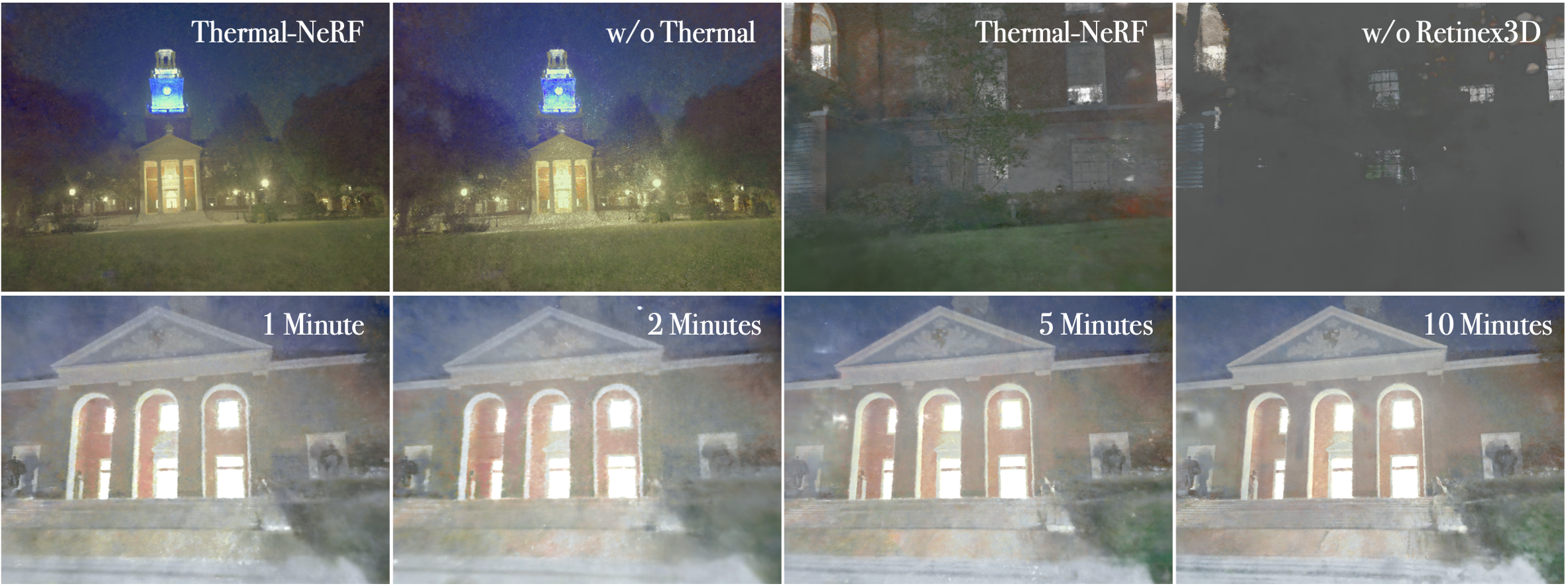}
\caption{Ablation study on thermal enhancement and Retinex3D strategies and demonstration of the training speed of our fast option Thermal-NeRF (iNGP).}
\label{fig:ablation}
\end{minipage}
\begin{minipage}[t]{0.48\textwidth}
    \includegraphics[width=\textwidth]{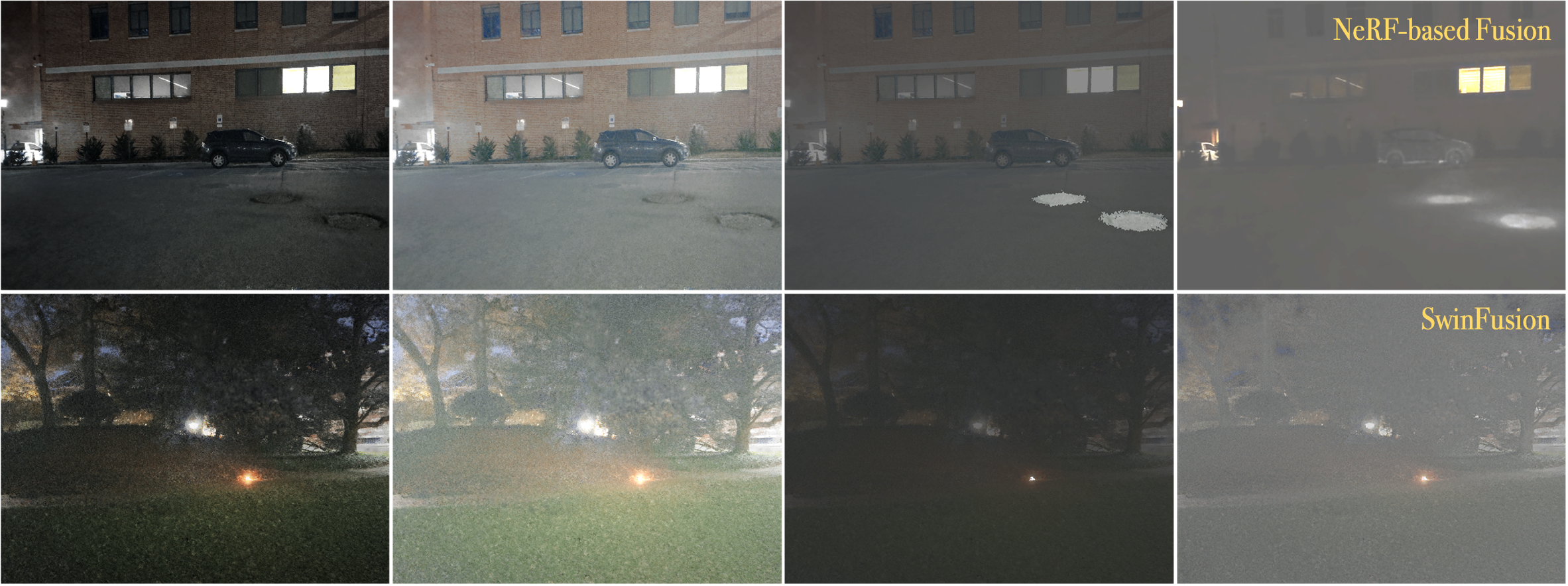}
\caption{From left to right, the images are generated by different post-processing, thermal object enhancement, and NeRF-based fusion or SwinFusion\cite{swinfusion}.}
\label{fig:app}
\end{minipage}
\end{figure}

Thermal enhancement strategy will constrain the geometry of the scene and regularize the balanced loss function, and Figure \ref{fig:ablation} shows that thermal enhancement is effective in the removal of the spot noises and floaters. Note that Multi-Exposure is established based on the Retinex3D strategy by fixing the reflectance field and modifying the illuminations. If we instead attach multiple independent RGB heads to the model, the color information is not shared between them. Also, there exists a saturation issue when scaling the linear raw images to generate pseudo exposures, so each head may present some white predictions when saturated, and the final estimation tends to be white. With the iNGP implementation, our model can present satisfactory reconstruction with 10 minutes of training on a single RTX3090, while RawNeRF (Mip-NeRF) requires around 22 hours of training to obtain high-quality results.

\begin{figure*}[t]
\centering
    \includegraphics[width=\linewidth]{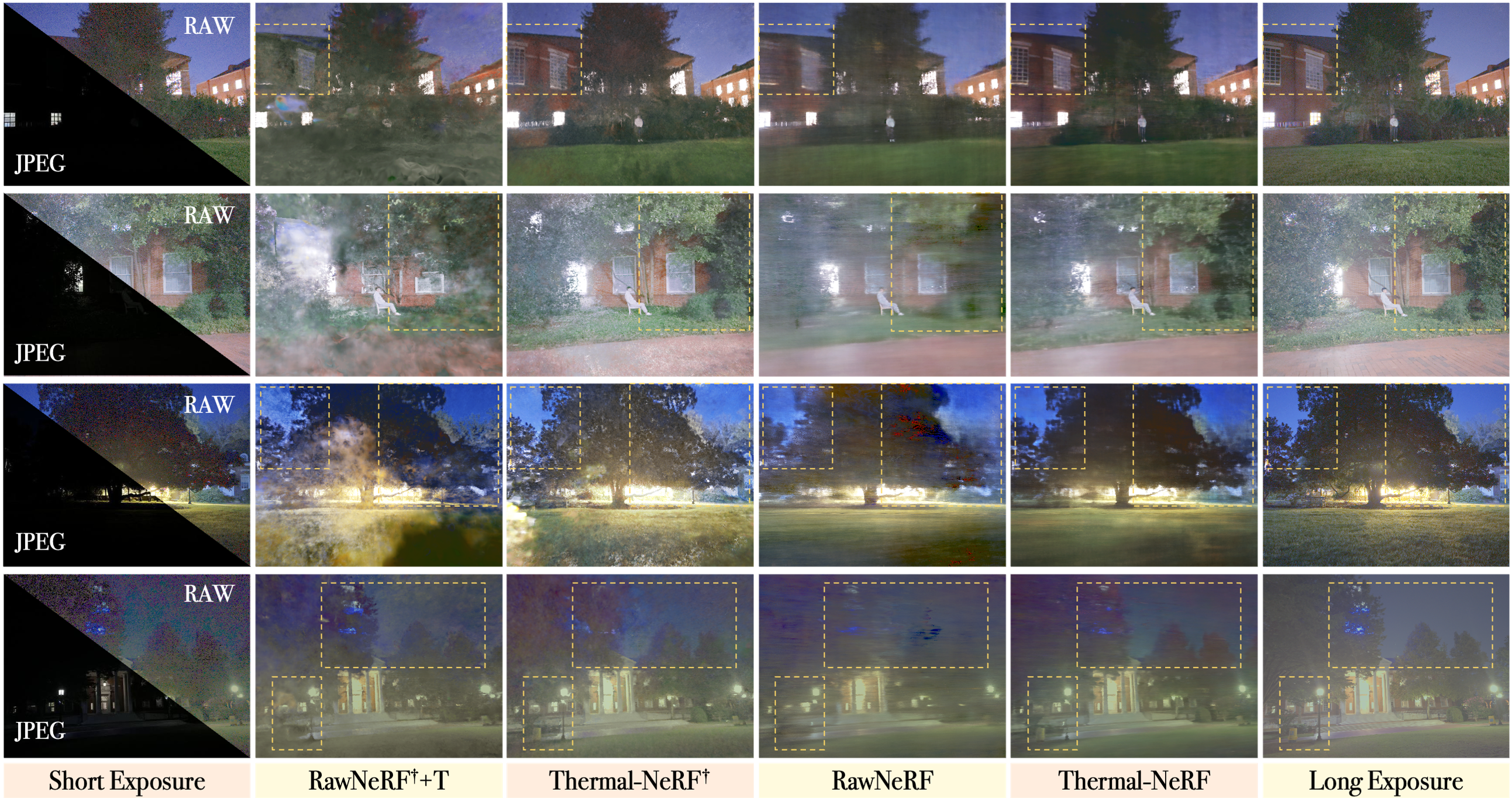}
\caption{Samples of experimental results on our MVTV dataset for visual comparison between different methods. The method notations follow Table \ref{tab:vs} for convenience. JPEG refers to the camera-finished images and indicates the darkness level of the scene. All the rendered raw images are post-processed by the same pipeline for visualization.}
\label{fig:vs}
\end{figure*}

Table \ref{tab:vs} presents the quantitative comparison between our model and RawNeRF. We omit the results for RawNeRF implemented on iNGP in Figure \ref{fig:vs} since it always converges to weird geometry and sometimes is unstable. Thermal enhancement will assist it in constraining the geometry, but its inherent issue that the single tiny MLP cannot model the highly imbalanced intensity distribution prevents it from better performance. As the statistics show, our fast model Thermal-NeRF\textsuperscript{\textdagger} presents similar results as RawNeRF with better reconstruction of the dark details (indicated by lower LPIPS), which originates from the multi-resolution gird explicit representation of iNGP. However, it shows weakness in noise smoothing in dark regions compared with Mip-NeRF based models. While RawNeRF is capable of noise removing, it also smooths the low-light texture and boundary details. Our slow model Thermal-NeRF achieves a better trade-off between detail preservation and noise smoothing and further improves the synthesis performance with the best PSNR scores. Notably, its rendered visual results approach long-exposure ground truth after post-processing.

\subsection{Extension}
%\noindent {\bf{Other Applications.}}
The high-quality and real-time rendition performance of 3DGS \cite{3dgaussiansplatting} inspired many improvement or adaptation works on different applications \cite{zhu2023fsgs, guedon2023sugar, gao2023relightable, chen2023neusg, yi2023gaussiandreamer}. However, 3DGS relies on controllable explicit geometry representation. The geometry information contained by the multi-view images is crucial for the convergence stability of 3DGS. In the supplementary part, we will show that directly moving the RawNeRF strategy to 3DGS will lead to unstructured Gaussian layouts, while our method can be seamlessly combined with 3DGS and present comparable synthesis performance alongside with high-speed rendering.

Given the rendered clean raw images, we can adjust the exposure level, gamma correction, tonemapping, and other post-processing steps to achieve a better visual effect on the visible environment (Figure \ref{fig:app}).
Also, the thermal objects can be enhanced by the rendered thermal images to indicate their 3D locations in the scene, which is beneficial to the rescue or surveillance \cite{rescue} requiring more visual information in low-light conditions. With the multi-view visible and thermal images in our dataset, it is possible to move the research of general 2D image fusion task \cite{fusiongan, ifcnn, swinfusion, res2fusion} to 3D, such as NeRF-based image fusion.

\section{Conclusion}
In this paper, we propose to leverage visible and thermal images together for novel view synthesis in extremely dark situations where only short-exposure images are available. Firstly, a multi-view thermal and visible dataset is established to support the research on multimodal NeRF. Then, we present Thermal-NeRF, which simultaneously accomplishes thermal and visible view synthesis and shows better reconstruction performance than previous work. Also, the proposed method can be seamlessly transferred to models with explicit representations for high-speed rendering. Finally, we demonstrate that both modalities are beneficial to each other in 3D low-light scene reconstruction.

\clearpage  % TODO REVIEW/FINAL: This \clearpage needs to be removed from both review and camera-ready versions.

% ---- Bibliography ----
%
% BibTeX users should specify bibliography style 'splncs04'.
% References will then be sorted and formatted in the correct style.
%
\bibliographystyle{splncs04}
\bibliography{main}

\title{Supplementary Materials: \\
Leveraging Thermal Modality to Enhance Reconstruction in Low-Light Conditions} 

% TODO REVIEW: If the paper title is too long for the running head, you can set
% an abbreviated paper title here. If not, comment out.
\titlerunning{Thermal-NeRF: Supplementary Materials}
\author{}
% \author{Jiacong Xu\inst{1} \and
% Mingqian Liao\inst{2} \and
% K Ram Prabhakar\inst{1} \and
% Vishal M. Patel\inst{1}}

% TODO FINAL: Replace with an abbreviated list of authors.
\authorrunning{Xu et al.}
% First names are abbreviated in the running head.
% If there are more than two authors, 'et al.' is used.

% TODO FINAL: Replace with your institution list.
\institute{}
% \institute{Johns Hopkins University, Baltimore MD 21218, USA \\
% \email{\{jxu155,rprabha3,vpatel36\}@jhu.edu} \and
% Southern University of Science and Technology, Shenzhen, P.R.China \\
% \email{12012919@mail.sustech.edu.cn}}

\maketitle

\section{MVTV}
\begin{table*}[ht]
\scriptsize
\centering
\begin{tabular}{c|cccccccccc|c} 
\hline\hline
VS       & Gilman & Shriver & Mason & Human & Road  & Chair & Corner & Tree & Parking & Forest & Average  \\ 
\hline
\#Images & 51     & 46      & 54    & 49    & 26    & 35    & 45     & 42   & 43      & 35     & 43       \\ 
\hline
TS       & Car0   & Car1    & Fire0 & Fire1 & Fire2 & Fire3 & Man0   & Man1 & Man2    & Bench  & Average  \\ 
\hline
\#Images & 45     & 44      & 43    & 42    & 39    & 41    & 49     & 44   & 56      & 42     & 45       \\
\hline\hline
\end{tabular}
\caption{Dataset statistics of VS and TS parts of MVTV. In addition to thermal and visible camera-finished images, VS part also provides long-exposure visible images as ground truth for visible synthesis in low-light conditions.}
\label{tab:mvtv}
\end{table*}
In this paper, we establish the first multi-view thermal and visible dataset (MVTV) for outdoor scenes in low-light conditions, and the corresponding statistics are presented in Table \ref{tab:mvtv}. MVTV contains over 40 forward-facing views for each scene on average. With the rigid transformation between two cameras, we only need to annotate one image for estimation of the homography matrix, and all the paired thermal and visible images can be aligned by this calculated matrix. Then, the overlapping area is clipped out for a standard image with a rectangular shape.  
Unlike advanced visible cameras, thermal cameras usually contain heavy distortion. Thus, we collect 30 images of a checkerboard under sunshine using the thermal camera and estimate the distortion coefficients to undistort the thermal images before alignment. The cameras are mounted on a tripod for stabilization during the long exposure (tens of seconds) to avoid blurring.

\section{Ablation Study}
\begin{figure}[ht]
\centering
    \includegraphics[width=\linewidth]{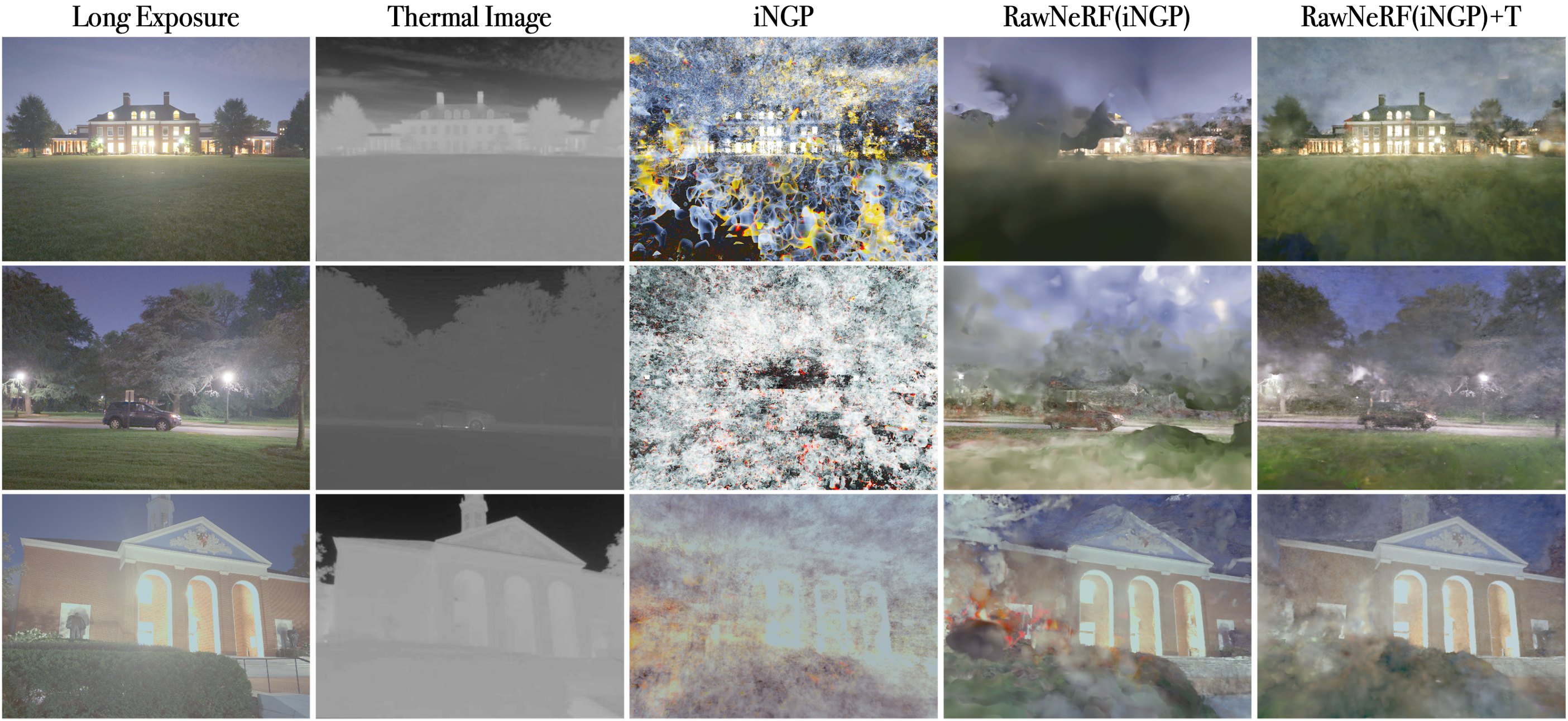}
\caption{Illustrations of the effectiveness of Thermal Enhancement (T) on RawNeRF \cite{rawnerf} implemented on iNGP \cite{instantngp}.}
\label{fig:thermal}
\end{figure}
\begin{figure}[ht]
\centering
    \includegraphics[width=\linewidth]{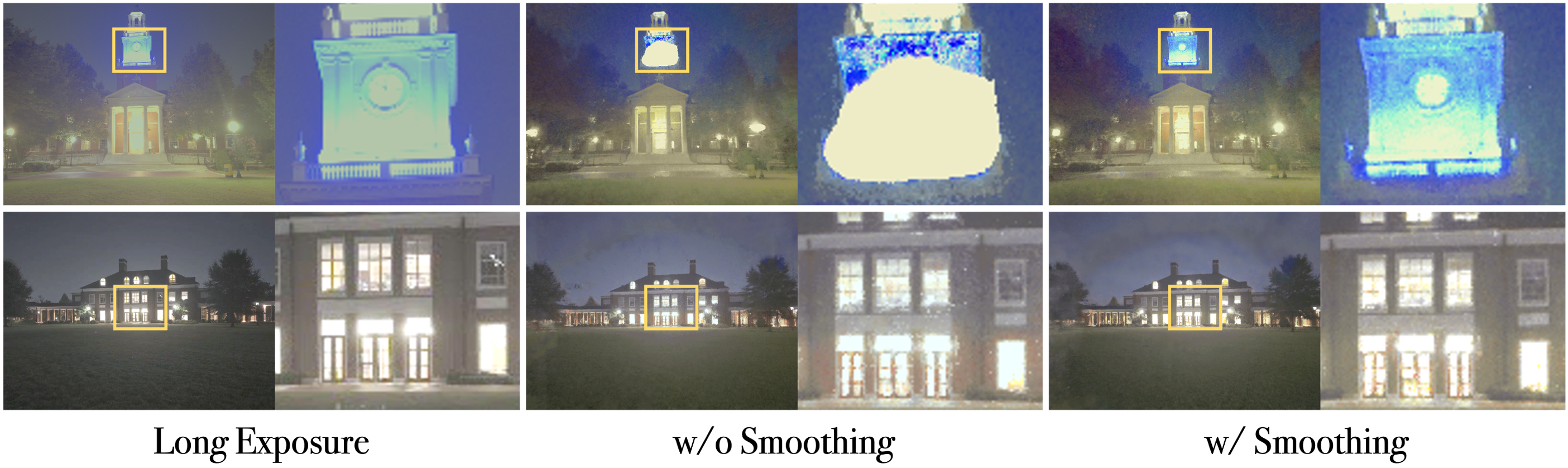}
\caption{Visualization of the results with and without the predefined smoothing sign function. Color inconsistency and spot points are observed in the middle images.}
\label{fig:sign}
\end{figure}
\begin{figure}[ht]
\centering
    \includegraphics[width=\linewidth]{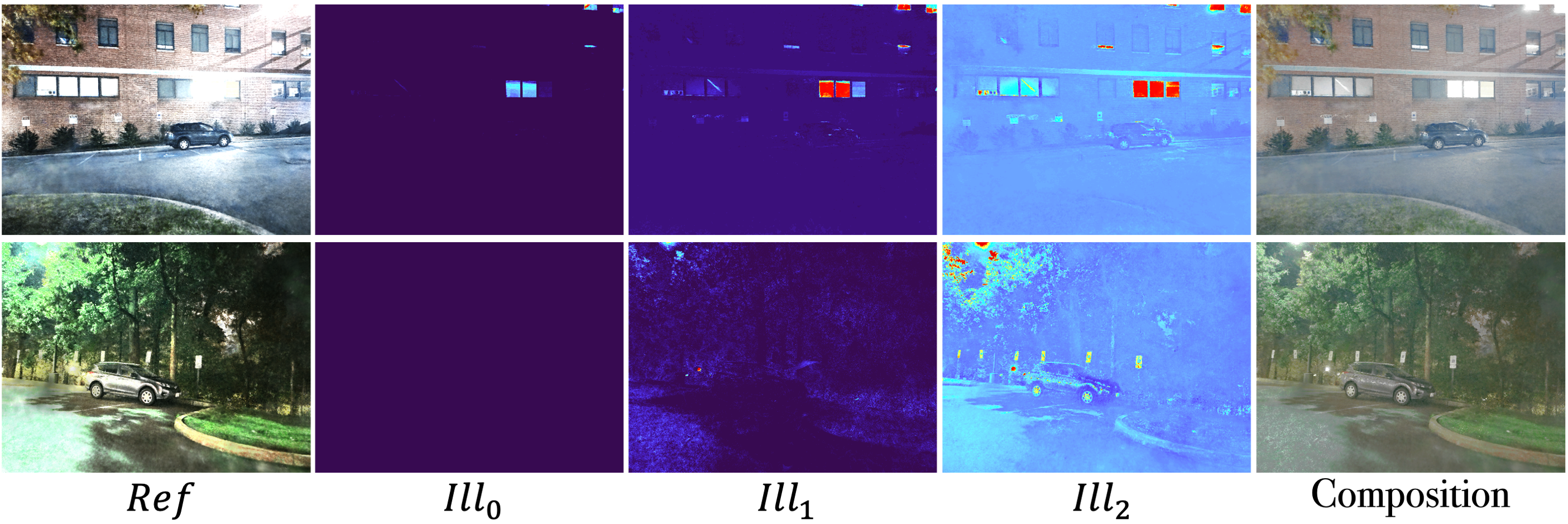}
\caption{Illustration of Retinex3D decomposition and illumination manipulation for Multi-Exposure strategy.}
\label{fig:me}
\end{figure}
\textit{i)} Optimizing the network primarily with imbalanced RGB data can cause the model to prioritize brightness reconstruction over geometric accuracy, as illustrated in Figure \ref{fig:thermal}. Thermal imagery, on the other hand, provides a comprehensive view of the scene's geometry. By integrating the thermal enhancement strategy, we can mitigate geometry-related errors, such as floaters, ensuring a more accurate representation of the structure. \textit{ii)} Given that the predictions from all the heads cannot be precise and include errors, employing a straightforward sign function would lead to color inconsistencies and the appearance of spot points in the ultimate reconstruction. Figure \ref{fig:sign} shows that our smoothing sign function successfully alleviates this issue. \textit{iii)} Adjusting illumination is more effective and intuitive than modifying reflectance because illumination involves only a single channel, and the color of each point in space remains constant regardless of changes in exposure. (Figure \ref{fig:me}).

\begin{figure}[ht]
\centering
    \includegraphics[width=\linewidth]{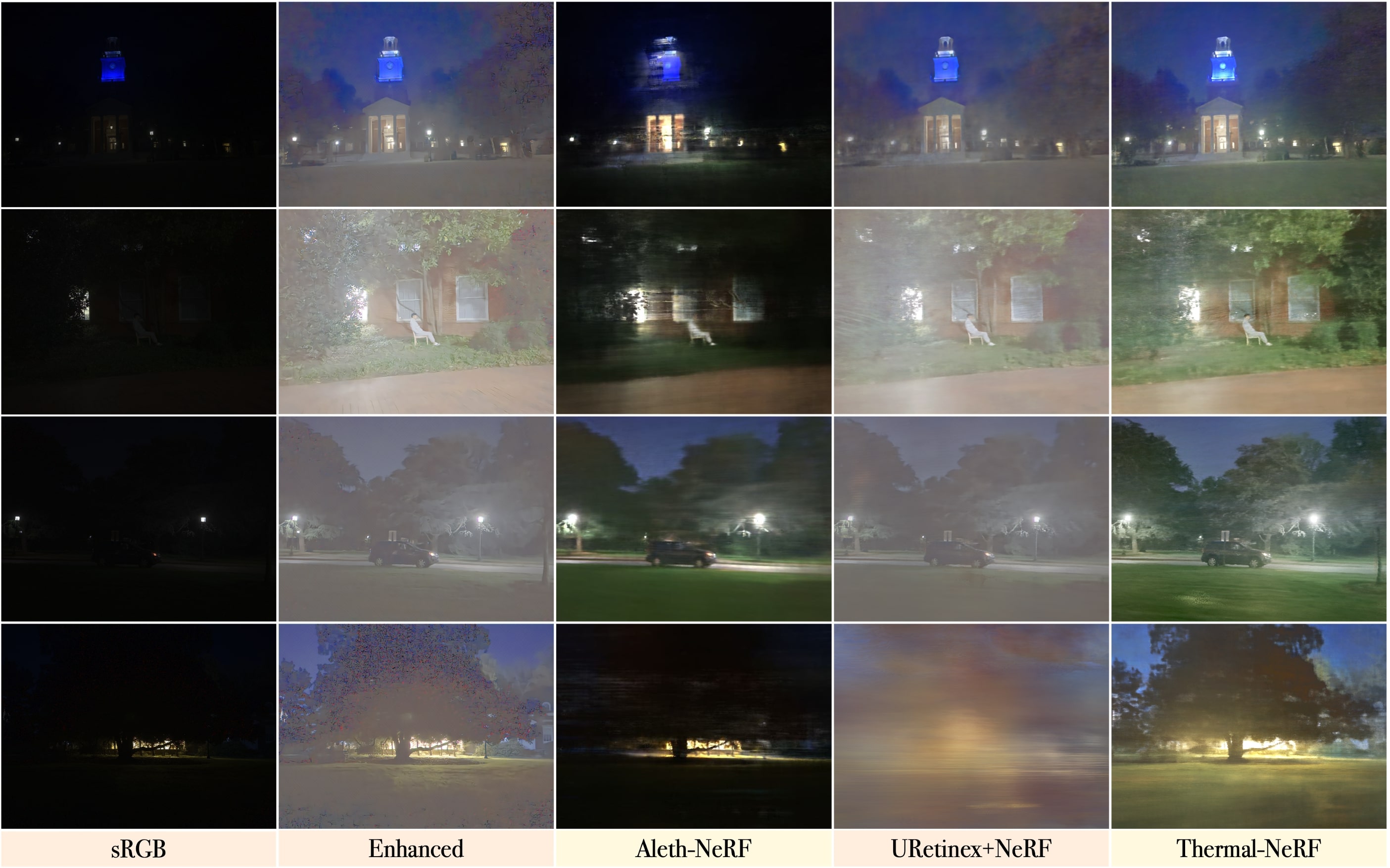}
\caption{Comparison between existing low-light NeRF methods \cite{alethnerf} (using sRGB as input) and Thermal-NeRF. Enhanced images are generated by URetinex \cite{uretinex} on sRGB.}
\label{fig:enhance}
\end{figure}
\section{sRGB vs RAW}
Different from RawNeRF \cite{rawnerf}, which reconstructs the dark scenes directly from raw images, Aleth-NeRF \cite{alethnerf} and LLNeRF \cite{llnerf} propose to accomplish view synthesis and image enhancement simultaneously using post-processed sRGB images. Note that raw images preserve the original camera sensor response in a large range of intensity (0 $\sim$ 65535), while sRGB images compress the dynamic range (0 $\sim$ 255) by a series of lossy post-processing steps, which correlate the noises and significantly reduce the colorimetric and geometric information in dark regions.

We apply existing methods based on sRGB images on our dataset, and the synthesis results are shown in Figure \ref{fig:enhance}. Since the codebase of LLNeRF \cite{llnerf} is not compatible with our system, we replace it with URetinex+NeRF, which enhances the training images by URetinex \cite{uretinex} before feeding the images to Mip-NeRF \cite{mipnerf} and presents similar reconstructions performance with LLNeRF as their paper states. The outputs of modern cameras are usually in JPEG format, which is a traditional lossy compression algorithm and results in blocks of zero values in the dark areas. To preserve the information as much as possible for a fair comparison with these methods, we generate sRGB images in PNG format using collected raw images following the default post-processing pipeline.

Figure \ref{fig:enhance} provides the comparison of synthesis results between Thermal-NeRF and existing sRGB-based methods. Aleth-NeRF \cite{alethnerf} learns an extra concealing field to model the darkness, so it is stable in the reconstruction of the scenes when the geometry information is insufficient. However, it shows weakness in the enhancement of the dark regions. URetinex+NeRF \cite{uretinex} presents better enhancement results, but it fails to recover the low-light geometry and texture details, distorts the noisy color signals (color inconsistency), and sometimes becomes unstable. These sRGB-based methods require many hyperparameters and another training process to adjust the brightness level, while Thermal-NeRF possesses more freedom with the rendered clean raw images on the post-processing pipeline. Therefore, we recommend reconstructing the highly dark scenes from raw images, which contain rich colorimetric information in the dark regions even though are full of noises, and the multi-view training process will integrate the information from different captures and serve as a multi-picture averaging method for noise reduction ($\sigma_{\bar y} \propto \sigma_{y}/\sqrt{N}$). 

% hyperparameters
% stablility
% geometry and color distortion

\section{Thermal-GS}
In this section, we will show that our method can be seamlessly applied in 3DGS \cite{3dgaussiansplatting} for real-time and high-quality rendering.
\subsection{3DGS Preliminary}
3D Gaussian Splatting (3DGS) \cite{3dgaussiansplatting} represents the scene explicitly by a collection of trainable 3D Gaussians. Each 3D Gaussian is characterized by its center $\mathbf{\mu} \in \mathbb{R}^{3}$ and 3D covariance matrix $ \mathbf{\Sigma} \in \mathbb{R}^{3\times 3}$ in world coordinates, and its influence on the opacity of a space point $\mathbf{x}$ is proportional to Gaussian distribution:
\begin{equation}
    G(\mathbf{x}) = e^{-\frac{1}{2}(\mathbf{x}-\mathbf{\mu})^{\top}\mathbf{\Sigma}^{-1}(\mathbf{x}-\mathbf{\mu})}
\end{equation}
Removing the 3D Gaussians outside the camera frustum, the remaining Gaussains are projected to the screen, called Splatting \cite{zwicker2001ewa}, for high-speed rendering (rasterization). Given the world-to-camera transformation matrix $\mathbf{W}$ and Jacobian of the affine approximation of perspective transformation $\mathbf{J}$, the projected 2D covariance matrix can be computed by:
\begin{equation}\mathbf{\Sigma}^{\prime}=\mathbf{J}\mathbf{W}\mathbf{\Sigma}\mathbf{W}^{\top}\mathbf{J}^{\top}
\end{equation}
Each 3D Gaussian stores the learnable shading information by opacity $\alpha_{i}$ and a set of spherical harmonic (SH) coefficients for view-dependent color $c_{i}$ computation. After sorting the Gaussians for every non-overlapping patch on the image by depth, alpha compositing is utilized to compute the final color for each pixel:
\begin{equation}
    C=\sum_{i=1}^{n}c_{i}\alpha_{i}^{\prime}\prod_{j=1}^{i-1}(1-\alpha_{j}^{\prime})
\end{equation}
where $\alpha_{i}^{\prime}$ is the multiplication of $\alpha_{i}$ and splatted 2D Gaussian $G^{\prime}(x^{\prime})$ , and $x^{\prime}$ is the coordinates in the projected space. Heuristic Point densification and pruning are employed in the training process for efficient 3D representation.

\begin{figure}[t]
\centering
    \includegraphics[width=\linewidth]{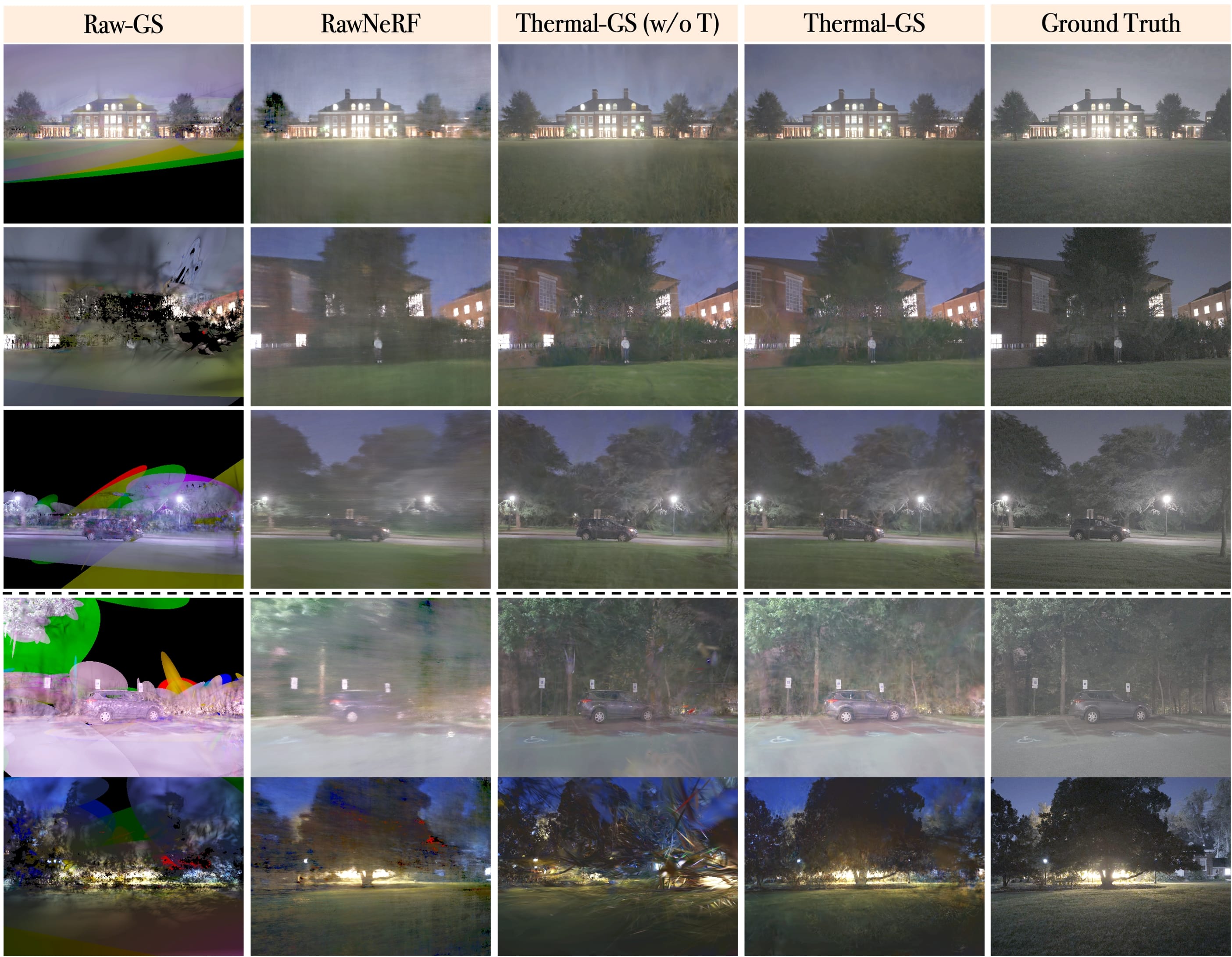}
\caption{Demonstration of the superiority of Thermal-GS over other methods. The last two rows present the failures of Thermal-GS without thermal modality (Thermal-GS (w/o T)), which only contains Retinex3D and Multi-Exposure. Raw-GS refers to 3DGS with the RawNeRF \cite{rawnerf} strategy.}
\label{fig:results}
\end{figure}
\subsection{Methodology Adaptation}
The original SH coefficients are preserved to describe the view-dependent reflectance $Ref$. Similar to iNGP \cite{instantngp} implementation, we attach 4 illumination heads for different scales of intensity prediction. Note that each set of SH coefficients contains 16 items, and a normal 3DGS requires millions of Gaussians to represent the scene. 4 view-dependent illumination heads will introduce a large amount of memory usage. Thus, we store a 16-dimension feature for each Gaussian and utilize a global Multi-Layer Perceptron (MLP) to predict illumination SHs, effectively halving the space complexity. 
\begin{equation}
    Ref=\sum_{i=1}^{n}\sigma(ref_{i})\alpha_{i}^{\prime}\prod_{j=1}^{i-1}(1-\alpha_{j}^{\prime}), \text{ } Ill^{k}=\sum_{i=1}^{n}ill^{k}_{i}\alpha_{i}^{\prime}\prod_{j=1}^{i-1}(1-\alpha_{j}^{\prime}),
\end{equation}
\begin{equation}
    \hat{C}^{k} = Ref \cdot Ill^{k}.
\end{equation}
where $k=0,1,2,3$ for multiple exposures, $\sigma(\cdot)$ refers to sigmoid function for constraining the reflectance value. Following the basic idea of our Retinex3D, the illuminations, and the reflectance are first rasterized to the canvas before Retinex multiplication, which will significantly reduce the noise. The Multi-Exposure strategy is employed to composite the final rendition.

\subsection{Experimental Results}
Training with raw images through the RawNeRF \cite{rawnerf} approach, 3DGS struggles to converge accurately on the correct geometry, leading to unrealistic results as shown in Figure \ref{fig:results}. Conversely, our approach, Thermal-GS, accomplishes visually plausible synthesis, outperforming in noise reduction and the preservation of details. Moreover, Figure \ref{fig:results} illustrates how the thermal enhancement strategy of our method stabilizes the reconstruction process by efficiently eliminating floaters. In the inference stage, the illuminations can be pre-computed from MLPs so that the rendering speed will not be affected.

\section{Video Results}
There are two attached video results: \textit{i)} Comparison of visible synthesis results between RawNeRF \cite{rawnerf} and Thermal-NeRF based on Mip-NeRF \cite{mipnerf} implementation. \textit{ii)} Demonstration of visible and thermal synthesis performance of Thermal-NeRF with iNGP \cite{instantngp} implementation and extended applications on object enhancement and image fusion (SwinFusion \cite{swinfusion}).

\section{Limitation}
Thermal cameras are sensitive to temperature variance, but when the temperature differences between objects are not significant, such as in indoor scenes (no human), thermal imaging fails to distinguish different objects. Besides, we do not observe extensive performance improvements in implicit NeRF models with thermal enhancement (usually only +0.3 PSNR). We argue that implicit models (Mip-NeRF \cite{mipnerf}) are not as sensitive to geometry information as explicit or hybrid models (iNGP \cite{instantngp} and 3DGS \cite{3dgaussiansplatting}), and implicit models inherently possess the capabilities of denoising and high dynamic modeling.

\end{document}